%% file: 2017-icra-popovic.tex
\documentclass[letterpaper, 10pt, conference]{ieeeconf/ieeeconf}    

\makeatletter
\let\NAT@parse\undefined
\makeatother

\usepackage[numbers,sectionbib,sort&compress]{natbib}

\usepackage{gensymb}
\usepackage{graphicx}
\usepackage{amsmath}
\usepackage{algorithm}
\usepackage[noend]{algpseudocode}
\usepackage{subcaption}
\usepackage[font=small]{caption}
\usepackage{sidecap} \sidecaptionvpos{figure}{c}
\DeclareMathOperator*{\argmax}{\arg\!\max}

\usepackage{hyperref}
\hypersetup{colorlinks,breaklinks,
linkcolor=[rgb]{0.5,0.,0.},
citecolor=[rgb]{0.000,0.427,0.173},
urlcolor=[rgb]{0.031,0.318,0.612}}

\usepackage[printonlyused,withpage,nolist,nohyperlinks]{acronym}
\input{common_acronyms.tex}

\providecommand{\figref}[1]{\mbox{Fig.~\ref{#1}}}
\providecommand{\eqnref}[1]{\mbox{Eq.~\ref{#1}}}

\renewcommand{\algref}[1]{\mbox{Alg.~\ref{#1}}}
\providecommand{\secref}[1]{\mbox{Section \ref{#1}}}

\IEEEoverridecommandlockouts                           
\title{\LARGE \bf
Online Informative Path Planning for Active Classification Using UAVs
}

\author{$\text{Marija Popovi\'{c}}^{1}$, $\text{Gregory Hitz}^{1}$, $\text{Juan Nieto}^{1}$, 
$\text{Inkyu Sa}^{1}$, $\text{Roland Siegwart}^{1}$, and $\text{Enric Galceran}$\thanks{${}^1$ 
Autonomous Systems Lab., Department of Mechanical and Process Engineering, ETH Z\"{u}rich, Z\"{u}rich, 
Switzerland. \texttt{mpopovic@ethz.ch}}%
}

\begin{document}

\maketitle
\thispagestyle{empty}
\pagestyle{empty}

\begin{abstract}

In this paper, we introduce an \ac{IPP} framework for active classification using
\acp{UAV}. Our algorithm uses a combination of global viewpoint selection and 
evolutionary optimization to refine the planned trajectory in continuous 3D
space while satisfying dynamic constraints. Our approach is evaluated on the application of weed detection 
for precision agriculture. We model the presence of weeds on farmland using an occupancy grid and generate
adaptive plans according to information-theoretic objectives, enabling the \ac{UAV} to gather data 
efficiently.  We validate our approach in simulation by comparing against existing methods,
and study the effects of different planning strategies. Our results show that the proposed algorithm builds 
maps with over 50\% lower entropy compared to traditional ``lawnmower'' coverage in the same amount of time. 
We demonstrate the planning scheme on a multirotor platform with different artificial
farmland set-ups.
\end{abstract}

\section{INTRODUCTION} \label{S:introduction}
Autonomous mobile systems are increasingly being used to collect information about the Earth and its
ecosystems~\cite{Dunbabin2012a}. In many applications,
including agriculture~\cite{Detweiler2015}, gas detection~\cite{Dunbabin2012a, Marchant2014}, and marine
biology~\cite{Hitz2015}, robots can provide high-resolution data capturing the spatial and
temporal dynamics of complex natural processes. Equipped with sensors, such devices are a flexible,
cost-efficient alternative to procedures based on manual sampling or static sensor
networks~\cite{Dunbabin2012a}. An open challenge, however, is planning paths for efficient data-gathering
given constraints on fuel, energy, or time.

In this work, we consider \ac{IPP} for a \ac{UAV} in agricultural monitoring. The objective is to survey a 
farmland using an on-board image-based weed classifier to quickly find precision treatment targets. By 
supplying crop health data required for targeted intervention, this workflow reduces chemical usage 
and yield loss, leading to sustainability and economic gain~\cite{Cardina1997}. In imaging, a key trade-off 
arises because the same point can be observed from different altitudes; thus, the planning unit must account 
for degrading sensor accuracy with increased altitude and coverage. Moreover, it must plan given limited 
battery and computational capacities.

We address the problem by proposing an \ac{IPP} framework for active classification in 3D space. We model the
presence of weed on farmland using an occupancy grid. We plan paths online through a combination
of global viewpoint selection and evolutionary optimization, which refines a continuous robot
trajectory while satisfying dynamic constraints. The resulting informative paths abide by a limited time
budget and address the challenge of trading off sensor resolution against coverage as discussed above.

\begin{figure}[h]
\centering
  \includegraphics[width=0.45\textwidth]{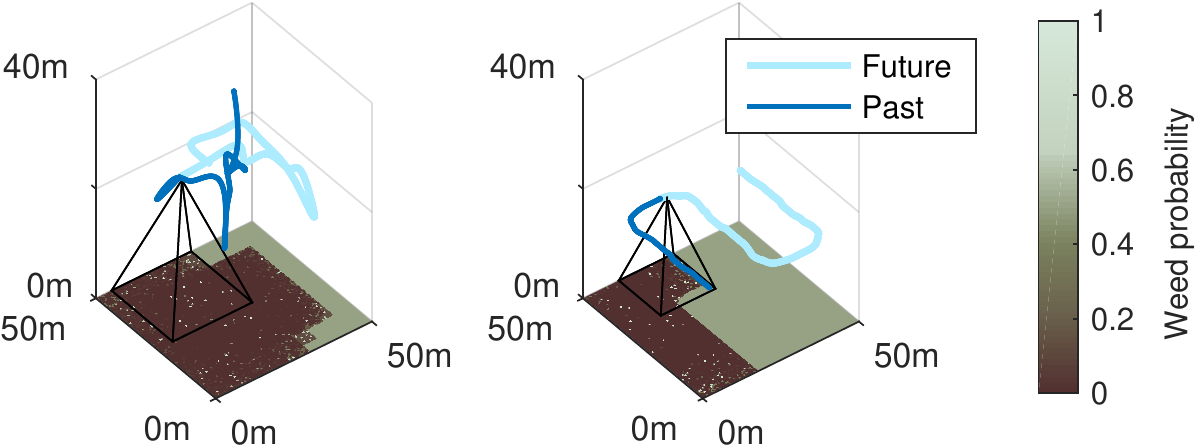}
    \vspace{5mm}

  \includegraphics[width=0.4\textwidth]{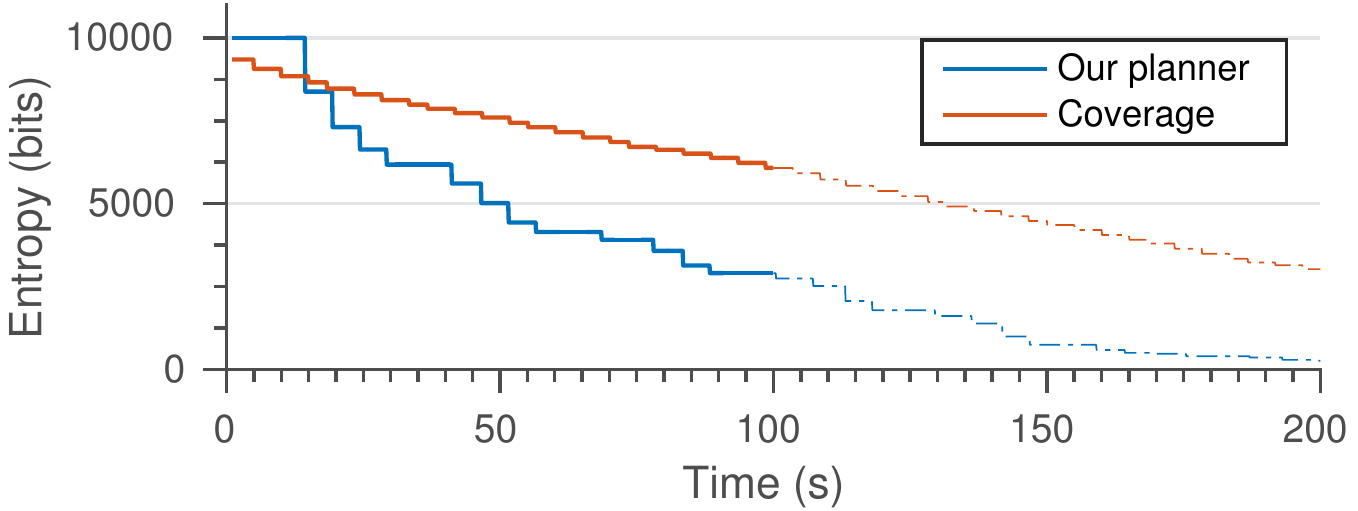}
  \vspace{0.1cm}
   \caption{A comparison of our IPP approach (top-left) to a ``lawnmower'' coverage path (top-right)
for an active weed classification task in precision agriculture using an \ac{UAV}. The pyramid shows the
camera footprint. The plot depicts the variations in map entropy over time. By planning adaptively with a
probabilistic sensor model, our approach produces a map with $45\%$ lower entropy of the coverage path
in the same amount of time (100s). }\label{F:teaser}
\end{figure}

The core contributions of this work are:
\begin{enumerate}
\raggedright
\item A new IPP algorithm with the following properties:
 \begin{itemize}
  \item generates dynamically feasible trajectories in continuous space,
  \item obeys budget and sensing constraints,
  \item uses a height-dependent noise model to capture sensor uncertainty.
 \end{itemize}
 \item The use of an evolutionary strategy to optimize continuous paths for maximum informativeness.
 \item A performance evaluation of our IPP algorithm in simulation against state-of-the-art planners and
a discussion of different planning strategies.
 \item Results from fully autonomously executed tests with AR tags as artificial weeds.
\end{enumerate}

\section{RELATED WORK} \label{S:related_work}
Significant work has been done recently on \ac{IPP} in robotics and related fields. In general, most
information-gain based strategies seek to minimize map uncertainty using objectives derived from Shannon's
entropy~\cite{Charrow2015a, Hollinger2014}. Unlike in distance-based planning, the path is subject to a
budget constraint limiting the number of measurements that can be taken. Formally, this problem can be
formulated as a \ac{POMDP}~\cite{Kaelbling1998}, which provides a general framework for uncertain planning.
However, the complexity of solving high-dimensional \ac{POMDP} models motivates more efficient solutions.

The NP-hard sensor placement problem~\citep{Krause2008} addresses selecting most informative measurement
sites in a static setting. Discrete \ac{IPP} algorithms build upon this task by performing combinatorial
optimization over a grid~\citep{Chekuri2005,Hollinger2009,Binney2013}. The main drawbacks of such
representations are their poor scalability and limited resolution. Alternatively, continuous-space planning
involves leveraging sampling-based methods~\citep{Hollinger2014} or
splines~\citep{Charrow2015a,Hitz2015,Marchant2014}. Our approach belongs to this class of methods, as it does
not require a predefined graph of viewpoint locations. Instead, similarly to~\citet{Charrow2015a}, we apply
global selection to identify promising viewpoints while escaping local minima, and optimization to refine our
trajectory in continuous 3D space.

We also distinguish between (i)~non-adaptive and (ii)~adaptive planning. Non-adaptive approaches explore an
environment using sequence of pre-determined actions. Adaptive approaches~\citep{Hitz2014, Lim2015,
Girdhar2015} exploit new measurements based on specific interests. Comparably to~\citet{Low2011}, we 
use a finite look-ahead, allowing us to propagate map changes and plan adaptively.

\ac{IPP} addressing \ac{UAV} imaging is a relatively unexplored area. Recently,~\citet{Vivaldini2016}
proposed a planner using Bayesian Optimization for mapping diseased trees. Like us, they consider
continuous-space plans for active classification with a probabilistic sensor providing aerial imagery.
However, their strategy is more computationally intensive as it requires interpolation of a pixel-based map.
Moreover, the planning algorithm does not allow for variable-altitude flight. The latter issue has been
addressed by~\citet{Sadat2015} in a similar set-up. Their method assumes discrete viewpoints and prior
knowledge of target regions, neglecting sensor noise. In contrast, our approach considers a height-dependent
sensor model and incrementally replans as data are collected. Furthermore, we use smooth polynomial
trajectories which guarantee feasibility under the \ac{UAV}'s dynamic constraints.

\section{PROBLEM STATEMENT} \label{S:problem_statement}
We define the general IPP problem as follows. We seek a continuous path $P$ in
the space of all possible paths $\Psi$ for maximum gain in some information-theoretic
measure:
\begin{equation}
\begin{aligned}
  P^* ={}& \underset{P \in \Psi}{\argmax}
\frac{I[\textsc{measure}(P)]}{\textsc{time}(P)}\textit{,} \\
 & \text{s.t. } \textsc{time}(P) \leq B \textit{,}
 \label{E:ipp_problem}
\end{aligned}
\end{equation}
where $B$ denotes a time budget and $I$ defines the utility function which
quantifies the informative objective. The function
\textsc{measure(\textperiodcentered)} obtains discrete measurements along the
path $P$ and \textsc{time(\textperiodcentered)} provides the corresponding
travel time. Maximizing the utility  \textit{rate} in~\eqnref{E:ipp_problem}
enables comparing the values of paths over different time scales, as opposed to
maximizing only the utility itself.

The above formulation uses a generic utility function $I$ to express the 
expected reduction in the map's uncertainty. In~\secref{S:background}, we consider Shannon's entropy 
and classification rate as possible informative measures for our application.

\section{BACKGROUND} \label{S:background}
We begin with brief descriptions of our approaches to parametrization as key concepts
underlying our \ac{IPP} algorithm.

\subsection{Environment and Measurement Models}
We represent the environment (a farmland above which the \ac{UAV} flies) using a 2D occupancy
grid $\mathcal{M}$~\citep{Elfes1989}, where each cell is associated with a Bernoulli random variable
indicating the probability of weed occupancy. For our measurement model, we assume a rectangular footprint
for a down-looking camera providing input to a weed classifier. The classifier provides weed occupancy for
cells within \ac{FoV} from a \ac{UAV} configuration $\textbf{x}$. For each observed cell
$\textbf{m}_i \in \mathcal{M}$ at time $t$, we perform a log-likelihood update given an observation $z$:
\begin{equation}
  L(\textbf{m}_i | z_{1:t}, \textbf{x}_{1:t}) =
L(\textbf{m}_i | z_{1:t-1}, \textbf{x}_{1:t-1}) +
L(\textbf{m}_i
| z_t, \textbf{x}_t) \text{,}
 \label{E:occupancy_grid_update}
\end{equation}
where $L(\textbf{m}_i | z_t, \textbf{x}_t)$ denotes the height-dependent sensor model capturing the weed
classifer output.

\begin{SCfigure}[][h]
  \includegraphics[width=0.258\textwidth]{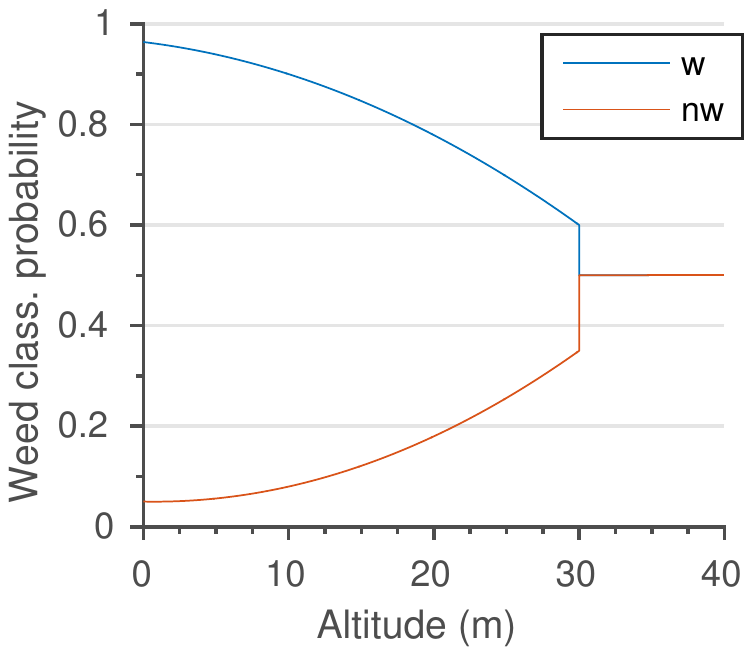}
   \caption{Probabilistic sensor model for a typical weed classifier. The blue and orange
curves depict the probability of label ``w'' given that ``w'' or ``nw'' was observed,
respectively. As altitude increases, the curves approach unknown classification
probability (0.5).}\label{F:sensor_model}
\end{SCfigure}

In our experiments, we use a binary weed classifier labeling observed cells as
``weed'' (w) or ``non-weed'' (nw). For each class, we define curves for our
sensor model~(\figref{F:sensor_model}) accounting for poorer classification with
high-altitude, low-resolution images. At low altitudes, our classifier
confidence levels match real datasets~\citep{Haug2014}, and we set a maximum
operating altitude, beyond which the classifier cannot provide any information.
To account for classifier processing times and limit the rate of information
gain in ~\eqnref{E:ipp_problem}, we also set a minimum time between consecutive
measurements.

\subsection{Path Parametrization}
To create paths abiding by the dynamic constraints of the \ac{UAV}, we connect
viewpoints $\textbf{x}\in\mathcal{X}$ using the method of~\citet{Richter2013}. As
in their work, we express a 12-degree polynomial trajectory in terms of
end-point derivatives, allowing for efficient optimization in
an unconstrained quadratic program.

\section{PATH PLANNING} \label{S:path_planning}
In this section, we present our \ac{IPP} framework. The main idea is to create fixed-horizon plans
maximizing an informative objective. To do this efficiently, we first select global viewpoints in 3D space
and then optimize the continuous path using an evolutionary method. We overview the steps of the
algorithm before discussing its key ingredients.

\subsection{Algorithm}
\begin{figure}[h]
\centering
\begin{subfigure}[]{0.14\textwidth}
  \includegraphics[width=\textwidth]{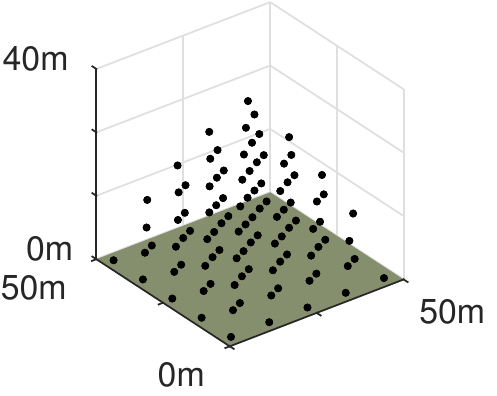}
  \vspace*{-5.5mm}
  \caption{}
  \label{SF:lattice}
     \end{subfigure}
\begin{subfigure}[]{0.1\textwidth}
  \includegraphics[width=\textwidth]{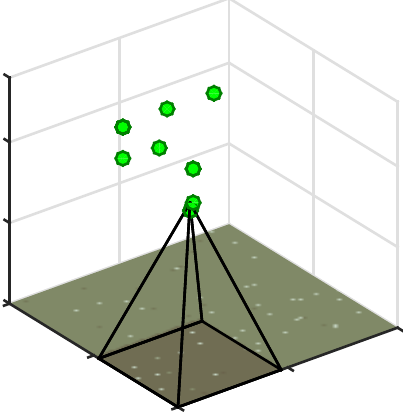}
  \caption{}
  \label{SF:global_points}
     \end{subfigure}
\begin{subfigure}[]{0.1\textwidth}
  \includegraphics[width=\textwidth]{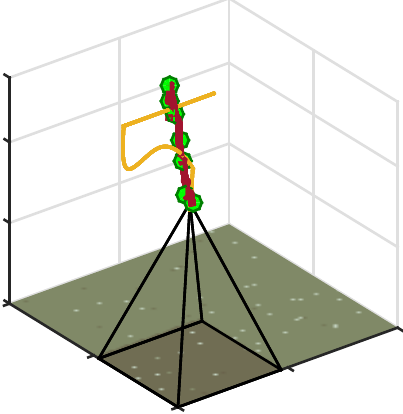}
    \caption{}
    \label{SF:global_optimization}
     \end{subfigure}
 \begin{subfigure}[]{0.1\textwidth}
  \includegraphics[width=\textwidth]{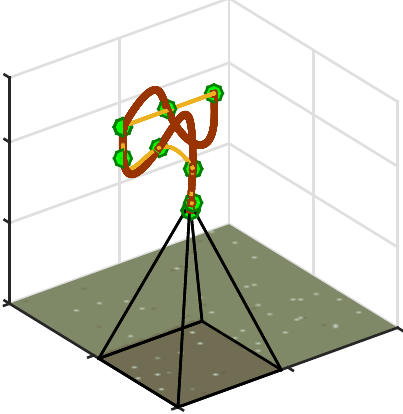}
    \caption{}
    \label{SF:intermediate_optimization}
     \end{subfigure}
\caption{(a) visualizes the lattice used by our planner. (b) shows the global viewpoints selected on the 
lattice. (c) and (d) depict subsequent global and local trajectory refinements. The orange and maroon curves 
show paths before and after optimization using the CMA-ES, respectively.}\label{F:replan_path}
\end{figure}

We use a fixed-horizon approach to plan adaptively. During the mission, we
maintain measurement viewpoints $\mathcal{X}$ within a horizon $H$, which is expressed in the number of 
points. We alternate plan execution and replanning, stopping when the elapsed time $t$ exceeds a budget 
$B$. We adopt a two-stage replanning approach consisting of global
viewpoint selection (Lines~3-10) and optimization (Line~11). This procedure is described
in~\algref{A:replan_path} and illustrated in~\figref{F:replan_path}. The following sub-sections detail the
key steps of~\algref{A:replan_path}.

\begin{algorithm}[h]
\renewcommand{\algorithmicrequire}{\textbf{Input:}}
\renewcommand{\algorithmicensure}{\textbf{Output:}}
\algrenewcommand\algorithmiccomment[2][\scriptsize]{{#1\hfill\(\triangleright\)
\textcolor[rgb]{0.4, 0.4, 0.4}{#2} }}
\begin{algorithmic}[1]

  \State $\mathcal{X}^g, \mathcal{X}^i \gets \emptyset$ \Comment{Initialize
global and intermediate viewpoints.}
  \While {$\textit{H} \geq |\mathcal{X}^g \cup \mathcal{X}^i|$}
   \If {$\textit{t/B} < \Call{rand}$ } \Comment{Select global objective based on time.}
   \State $\textbf{x}^* \gets$ Select viewpoint in $\mathcal{L}$
using~\eqnref{E:info_objective}
  \Else
    \State $\textbf{x}^* \gets$ Select viewpoint in $\mathcal{L}$
using~\eqnref{E:class_objective}
  \EndIf
  \State $\mathcal{M} \gets$ \Call{simulate\_measurement}{$\mathcal{M}$,
$\textbf{x}$} \Comment{Using ML.}
  \State $\textit{t} \gets \textit{t}$ + \Call{time}{$\textbf{x}^*$}
  \State $\mathcal{X}^g \gets \mathcal{X}^g \cup \textbf{x}^*$
  \State $\mathcal{X}^i \gets \mathcal{X}^i \cup \Call{add\_intermediate\_points}{\textbf{x}^*}$
  \EndWhile
  \State $\mathcal{X} \gets \mathcal{X}^g \cup \mathcal{X}^i$; $\mathcal{X} \gets$
\Call{cmaes}{$\mathcal{X}$, $\mathcal{M}$}
\Comment{Optimize polynomial.}

\end{algorithmic}
\caption{\textsc{replan\_path} procedure}\label{A:replan_path}
\end{algorithm}

\subsection{Global Viewpoint Selection}

In the first step (Lines~3-10), we sequentially select global measurement sites $\mathcal{X}^g$
(\figref{SF:global_points}). Unlike in frontier-based exploration common for indoor
mapping~\citep{Charrow2015a}, choosing viewpoints using map
boundaries is not applicable in our set-up. Instead, we greedily
apply~\eqnref{E:ipp_problem} over the horizon $H$ (Line~2) to
find most informative measurement sites. To find the next viewpoint
$\textbf{x}^*$ efficiently, we evaluate the objective over a multiresolution
lattice $\mathcal{L}$ (\figref{SF:lattice}). To encourage exploration,
we maximize entropy reduction in $\mathcal{M}$:
\begin{equation}
  I  [t+1|t] = H(\mathcal{M}_{t}) - H(\mathcal{M}_{t+1}) \text{.}
 \label{E:info_objective}
\end{equation}
To encourage classification, we divide $\mathcal{M}$ into
``weed'' and ``non-weed'' cells using thresholds $\delta_{w}$ and $\delta_{nw}$, leaving an unclassified
subset $\mathcal{U} = \{m_i \in \mathcal{M}\,|\,\delta_{nw} < p(m_i) < \delta_w\}$. This
is similar to finding unknown space in conventional occupancy mapping. We maximize the reduction of
$\mathcal{U}$ between time-steps:
\begin{equation}
  I[t+1|t] = |\mathcal{U}_{t}| - |\mathcal{U}_{t+1}| \text{.}
 \label{E:class_objective}
\end{equation}
We include an optional time-varying parameter $t/B$ (Line~3) to gradually bias 
viewpoint selection towards~\eqnref{E:class_objective} from~\eqnref{E:info_objective}, focusing on weed 
identification over time. We then simulate a \ac{ML} measurement at $\textbf{x}^*$ (Line~7) and interpolate 
intermediate viewpoints $\mathcal{X}^i$ (Line~10) to add degrees of freedom to the polynomial path for 
optimization.

\subsection{Optimization}
In the second step (Line~11), we optimize the polynomial path by solving~\eqnref{E:ipp_problem} 
in~\secref{S:problem_statement} using the Covariance Matrix Adaptation Evolution Strategy 
(CMA-ES). We opt for this method as our discrete measurement model does not provide the 
continuity necessary for information gradient-based optimization. Moreover, the CMA-ES has been applied 
successfully for continuous curve fitting in constrained spaces matching our problem 
set-up~\citep{Hansen2006}.

In \secref{S:experimental_results}, we consider global viewpoint objectives for (i)~information
gain only (\eqnref{E:info_objective}), (ii)~classification gain only (\eqnref{E:class_objective}), and 
(iii)~using the time-varying parameter (\algref{A:replan_path}). For the CMA-ES, we consider (i)~globally 
optimizing $\mathcal{X}$ (\figref{SF:global_optimization}) and (ii)~optimizing $\mathcal{X}^i$ only
(\figref{SF:intermediate_optimization}) for inter-segment refinements. We refer to these two
optimization methods as the ``global'' and ``local'' CMA-ES, respectively. For the global CMA-ES,
the points in $\mathcal{X}^g$ vote on the optimization objective for the entire trajectory.

\section{EXPERIMENTAL RESULTS} \label{S:experimental_results}
In this section, we first evaluate our proposed \ac{IPP} framework in simulation by comparing it to existing
algorithms and study different the variants of our algorithm introduced in~\secref{S:path_planning}.
Then, we implement our complete system in an environment with artificial weed distributions.

\subsection{Comparison Against Benchmarks}
We validate our framework in simulation on 100 $50~\times~50$\,m farmland environments with randomly 
scattered 
weeds. To analyze how our algorithm behaves with different weed densities, we generate Poisson distributions 
with 50 to 250 weeds. We use a resolution of $0.5$\,m with thresholds of $\delta_{nw}~=~0.25$ and 
$\delta_{w}~=~0.75$ for the occupancy grid map. To simulate false measurements, uniform noise is added based 
on the same 
probabilistic distribution as our sensor model (\figref{F:sensor_model}). The number of false positive cells 
is limited to 800 to avoid excessive noise in non-occupied regions.

Our methods are evaluated against traditional ``lawnmower'' coverage and the sampling-based rapidly exploring
information gathering tree (RIG-tree) introduced by~\citet{Hollinger2014}, a state-of-the-art \ac{IPP}
algorithm. We specify a $300$\,s budget $B$. For the weed classifier, we set a $60\degree$ camera \ac{FoV}
with a square footprint and maximum measurement frequency of $0.2$\,Hz. Map entropy, classification rate, and
mean F2-score are considered as metrics common for classification tasks. Following a similar approach
to~\citet{Pomerleau2013}, the \ac{CDF} of entropy is computed over a time
histogram to summarize the variability among trajectories. For this metric, faster-rising curves 
represent quicker reductions in map uncertainty and thus better performances. We use mean F2-score as the 
accuracy statistic to emphasize the effects of relatively fewer false negative misclassifications.

The \ac{UAV} position for both \ac{IPP} schemes is initialized to the map center with $40$\,m altitude. The
reference velocity and acceleration for trajectory optimization are $3$\,m$/$s and $1.5$\,m$/$s$^2$. For our
planner, we use a replanning horizon $H$ of 5 viewpoints to limit optimization complexity. For
RIG-tree, we associate the cost of a vertex (viewpoint) with accumulated travel time, and its
information value as map entropy given a new measurement. To compute cost, trajectory
optimization is performed for each edge, assuming measurements taken from rest. As the map cells are
independent, we apply the modular pruning strategy described in~\cite{Hollinger2014}.

We provide RIG-tree with prior knowledge from a high-altitude scan for initial planning. Then, we alternate
between tree construction and plan execution to allow for adaptivity. Each tree construction is terminated
after the same $\sim20$\,s allowed for trajectory optimization in our planner.

For the coverage planner, we define a height ($14.43$\,m) and maximum velocity ($0.844$\,m$/$s) for complete
coverage given the specified budget. To provide a fair comparison to the \ac{IPP} planners, several coverage 
patterns were evaluated and the best-performing one selected.

\begin{figure}[h]
\centering
  \includegraphics[width=0.5\textwidth]{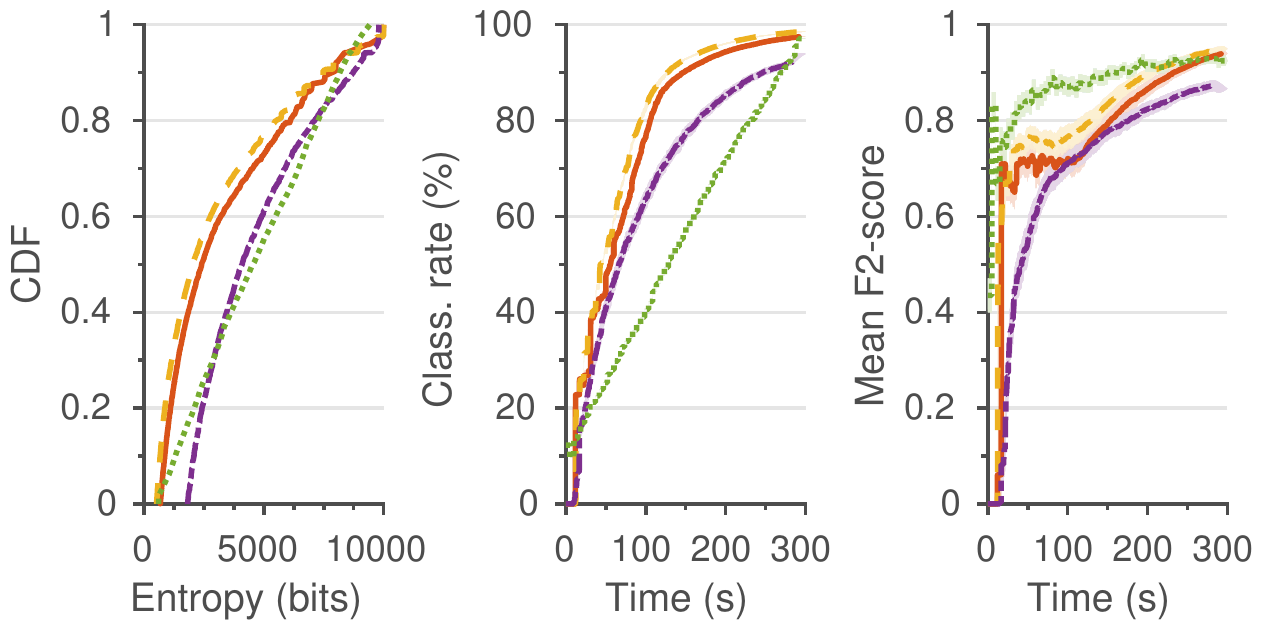}
   \vspace{-2.5mm}

    \includegraphics[width=0.48\textwidth]{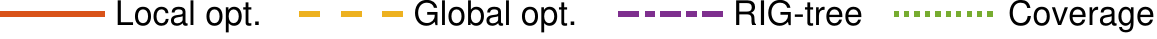}
   \caption{Comparison of our algorithms with RIG-tree and coverage path planners. For visualization, zero
planning time is assumed. The solid lines indicate means over 100 trials. The thin shaded regions depict 90\%
confidence bounds. Using \ac{IPP}, the informative metrics improve quickly as the \ac{UAV} flies at variable 
altitudes. The sharper curve slopes demonstrate that our algorithms perform better than RIG-tree
given the same set-up and planning time.}\label{F:evaluation}
\end{figure}

\figref{F:evaluation} shows how the algorithms score on the metrics during the mission. For our
planners, results using the time-varying objective (as in~\algref{A:replan_path}) are included since we found
it to be the most effective global viewpoint selection strategy. As expected, entropy reduction rates (left) 
are constant with na\"ive coverage (green) as the environment is scanned uniformly. In contrast, the \ac{IPP} 
methods perform better as they permit variable-altitude flight for wider \acp{FoV}. As shown 
in~\figref{F:teaser}, our planners usually produce paths resembling spirals, starting with descent to 
the unknown map center. Such motions permit the collection of low-quality, high-altitude data before focusing
on map corners and detected areas of interest.

The F2-score variations (right) suggest that coverage planning yields most accurate classification in 
the observed areas. This likely occurs due to the low flight altitude permitted by the 
allocated budget, basing noise additions on relatively certain regions of the sensor curves. We note 
that this metric does not account for the fact that, early in the mission, a large section of the map is 
completely unknown (\figref{F:teaser}).

Our planners (red, yellow) produce more informative paths than RIG-tree (purple) given the same 
planning time. This indicates that our strategy finds promising viewpoints in 3D space more efficiently 
than incremental sampling-based techniques. Moreover, it does not require prior knowledge for initialization 
and generates smooth trajectories.

For our planners, we observe that the global CMA-ES optimizer (yellow) produces both higher gains in entropy 
and classification rate compared to local optimization of intermediate points only (red). The following 
sub-section offers a more detailed comparison.

\subsection{Evaluation of Planning Strategies}

Next, we study changing planning strategies in our framework for the same simulation set-up to assess 
their effects. In our experiments, we consider varying:

\begin{itemize}
 \item \textit{Global viewpoint objectives}: information only (\eqnref{E:info_objective}), classification only
(\eqnref{E:class_objective}), using time-varying parameter (\algref{A:replan_path})
 \item \textit{Optimization methods}: no CMA-ES, local CMA-ES (\figref{SF:intermediate_optimization}), global
CMA-ES (\figref{SF:global_optimization})
\end{itemize}

\begin{figure}[h]
\centering
  \includegraphics[width=0.5\textwidth]{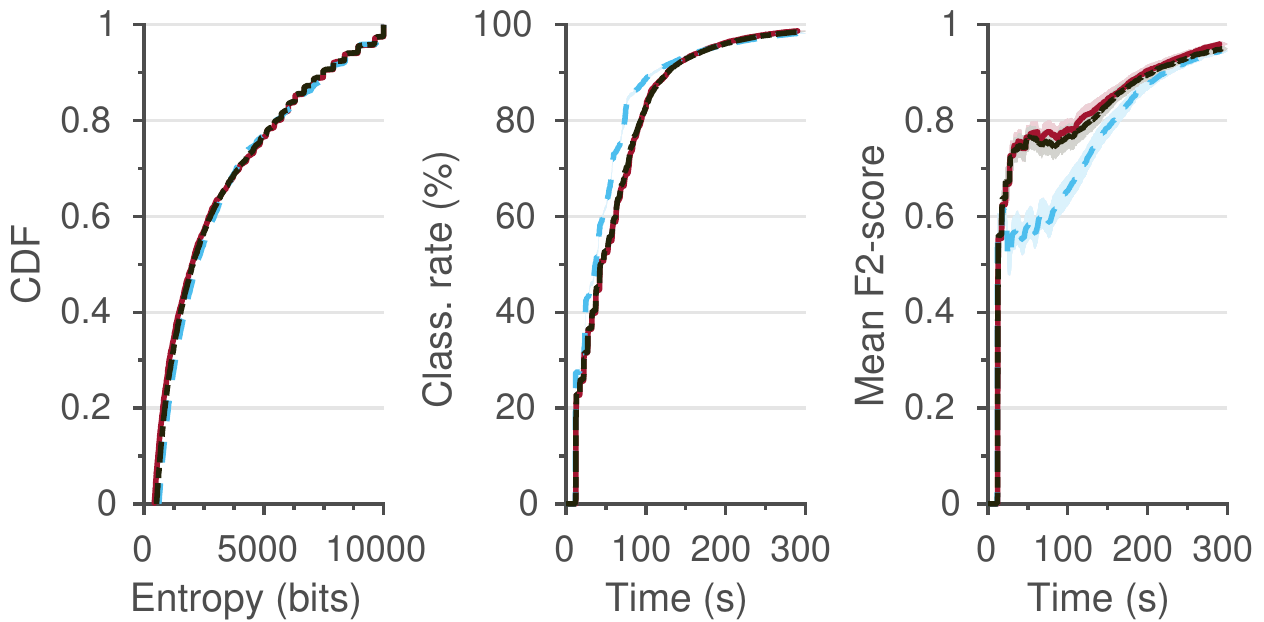}
  \vspace{-2.5mm}

    \includegraphics[width=0.42\textwidth]{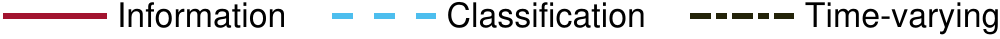}
   \caption{Comparison of greedy global viewpoint selection objectives on the multiresolution lattice. All
variants use global CMA-ES optimization. Accounting for spent budget (time) during the mission trades off the
information and classification objectives to provide a low-entropy map with high classification
rate.}\label{F:objectives}
\end{figure}

\figref{F:objectives} compares the global viewpoint selection objectives with the global CMA-ES. 
The curves illustrate the coverage-resolution trade-off: for the classification objective (light blue), 
flying at low altitudes quickly produces a map with cells within occupancy thresholds, as shown by the 
sharpest rise in classification rate (center). However, entropy reduction (left) is limited and initial 
accuracy (right) is poor since improving confidence on ``weed'' and ``nonweed'' labels is not accounted for. 
By considering elapsed time when selecting global viewpoints (black), we balance between improving existing 
information quality and exploring unknown areas to obtain a high certainty map with efficient classification.

\begin{figure}[h]
\centering
  \includegraphics[width=0.5\textwidth]{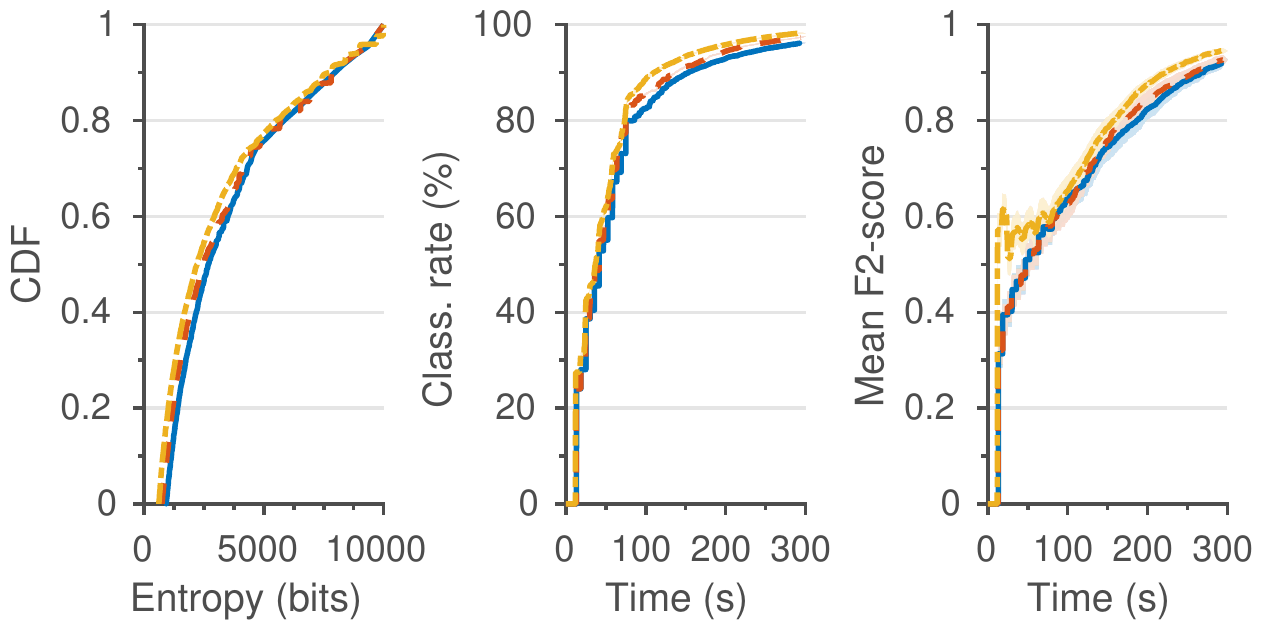}
  \vspace{-2.5mm}

    \includegraphics[width=0.38\textwidth]{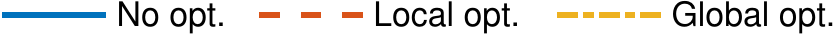}
   \caption{Comparison of CMA-ES optimization methods. All variants use the
classification objective for global viewpoint selection. The effect of local optimization
is marginal due to the small number of refined intermediate points on the planned trajectory.
Overall, global optimization performs best.}\label{F:optimizers}
 \end{figure}

\figref{F:optimizers} compares the CMA-ES optimization methods for the classification global viewpoint
selection objective. Optimizing the entire trajectory using the global CMA-ES 
(yellow) leads to best performance on all three metrics, likely due to the highest number of optimized 
variables. This highlights the effectiveness of using greedy selection to initialize an evolutionary-based 
trajectory optimizer, even in noisy conditions. In contrast, applying local optimization (red) on the 
short replanning horizon lead to minor improvements.

\subsection{Experiments}
\begin{figure}[h]
\centering
 \begin{subfigure}[]{0.31\textwidth}
  \includegraphics[width=\textwidth]{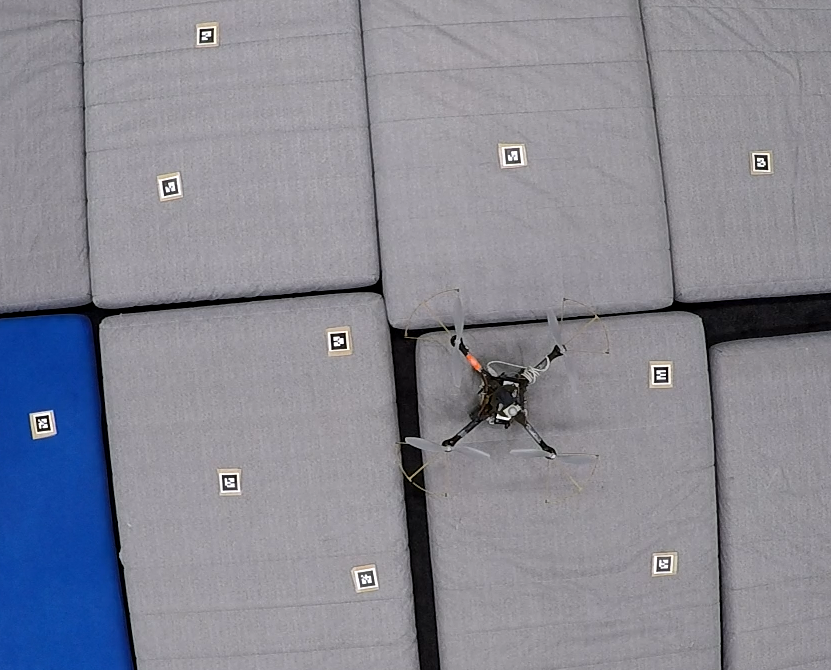} 
   \caption{} \label{SF:vicon_setup}
 \end{subfigure}
  \begin{subfigure}[]{0.16\textwidth}
  \includegraphics[width=\textwidth]{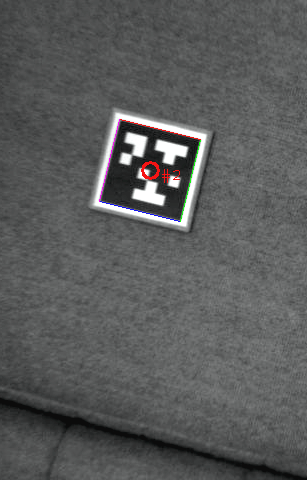} 
   \caption{} \label{SF:tag_detection}
 \end{subfigure}
\caption{(a) shows a top-down view of our experimental set-up with AR tags as simulated weeds. The 
red circle in (b) exemplifies an AprilTag detection output overlaid on part of a camera image. When 
taking a measurement, the detected tag pose is projected onto a pre-defined occupancy grid for informative 
planning.} 
\label{SF:environment}
\end{figure}

\begin{figure}[h]
\centering
  \includegraphics[width=0.5\textwidth]{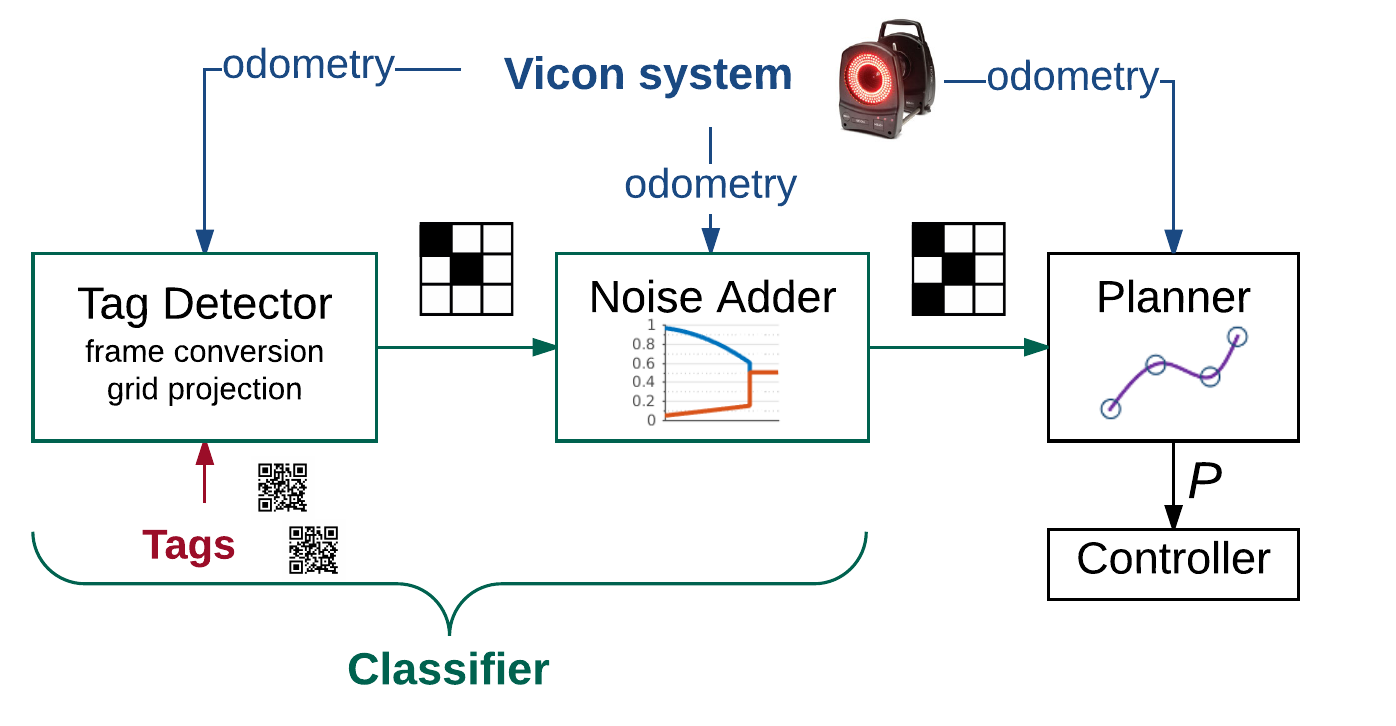}
   \caption{System diagram for our indoor experiments. An AR tag detector with optional added noise emulates
the weed classifier unit. Localization is provided by the Vicon system. The planner passes
fixed-horizon polynomial paths, $P$, to a model-predictive controller.}\label{F:experiment_setup}
\end{figure}

We show our \ac{IPP} strategy running in real-time on an AscTec Pelican \ac{UAV} platform. The 
experiments are conducted in an empty $4~\times~4$\,m indoor environment with a maximum altitude of $3$\,m, 
and state estimation provided by the Vicon motion capture system (\figref{SF:vicon_setup}). We 
use $6.4~\times~6.4$\,cm AR ground tags to mimic weeds. To emulate the weed classifier, we use a 
downward-facing FMVU-03MTM-CS 0.3MP Point Grey camera with the AprilTag library to detect the tags 
(\figref{SF:tag_detection}) given a $0.4$\,Hz maximum measurement frequency. The tags are then 
projected onto a $0.15$\,m resolution occupancy grid in the Vicon reference frame with thresholds of 
$\delta_{nw}~=~0.25$ and $\delta_{w}~=~0.75$. For a region observed from a particular height, probabilistic 
map updates are performed with a scaled version of the sensor model in~\figref{F:sensor_model}. As in the 
simulations, uniform noise can be generated on the emulated classifier output. In practice, however, we found 
the raw tag detection output uncertain enough so that additional noise was 
unneeded. ~\figref{F:experiment_setup} depicts the various modules in our set-up. The tag detector and 
Vicon systems represent generic units that can be replaced with a real online weed classifier and outdoor 
state estimation in field trials.

The aim is to show our algorithm operating with different real-life tag (weed) distributions. We consider 
scenarios of (i)~10 randomly distributed tags, (ii)~14 tags grouped in a cluster, and (iii)~no tags. In each 
case, we set the budget $B$ to $150$\,s the first measurement viewpoint to $2$\,m altitude 
at the map center, and the reference speed and acceleration to $0.5$\,m$/$s and $1.5$\,m$/$s$^2$. For 
planning, the information objective for global viewpoint selection is used with 
the global CMA-ES to demonstrate all elements of our strategy.

\begin{figure}[h]
\centering
  \includegraphics[width=0.46\textwidth]{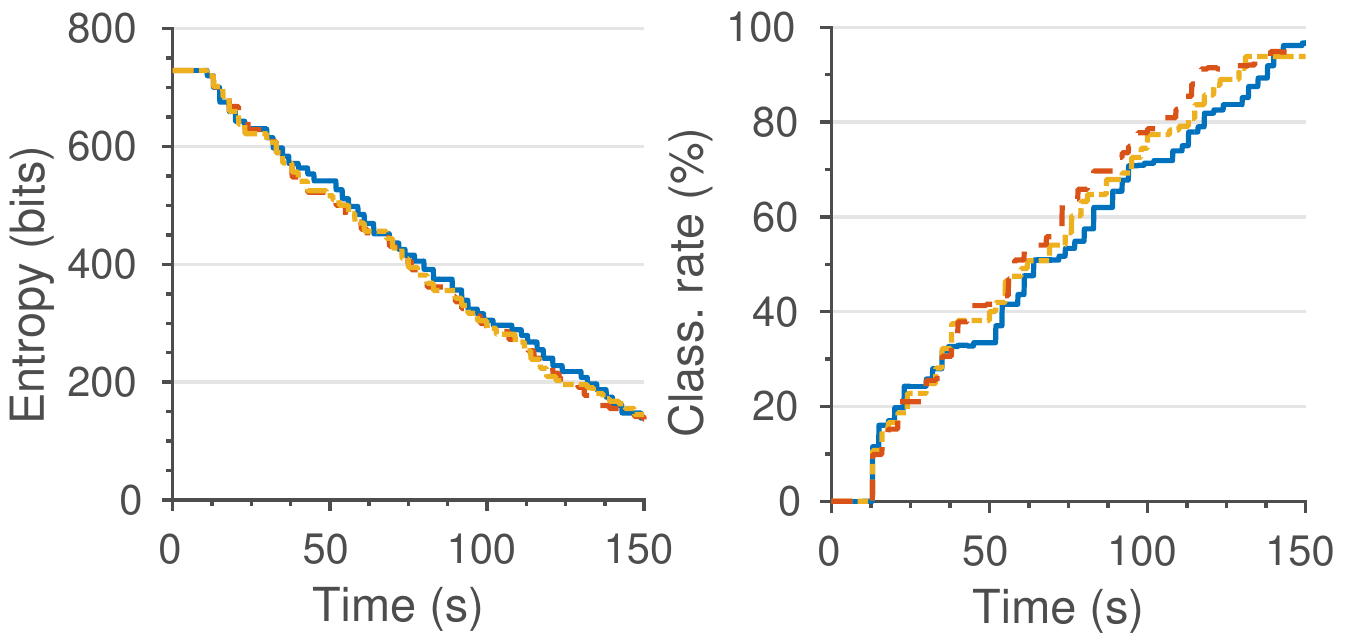}
  \vspace{1.05mm}
  
    \includegraphics[width=0.41\textwidth]{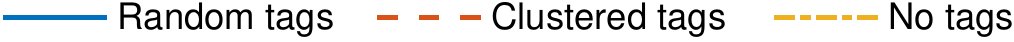}
   \caption{Informative metrics over time for our \ac{IPP} algorithm with different tag distributions and 
planning time taken into account. The planning strategy uses the information global viewpoint selection 
objective and the global CMA-ES. The expected reductions in map uncertainty validate our approach in 
real-time.}\label{F:experiment_metrics}
\end{figure}

\begin{figure}[h]
\centering
  \includegraphics[height=0.225\textwidth]{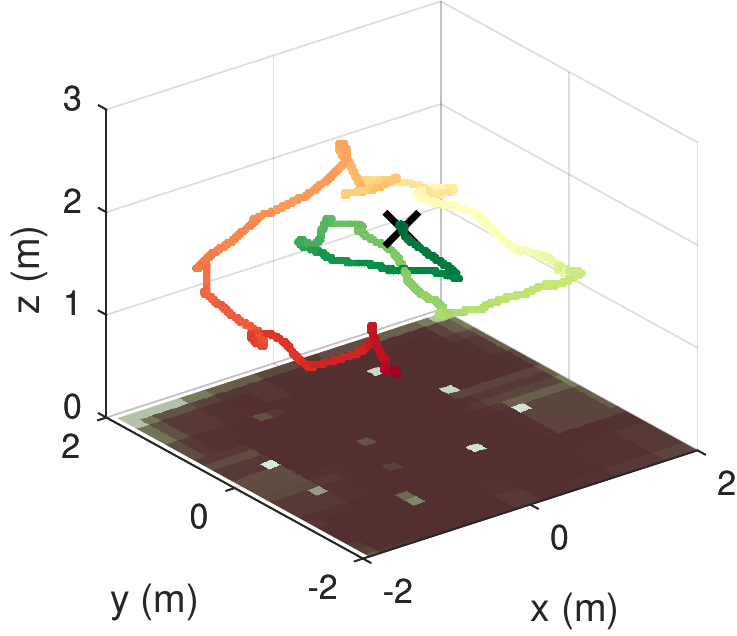}
  \hspace{6mm}
   \includegraphics[height=0.225\textwidth]{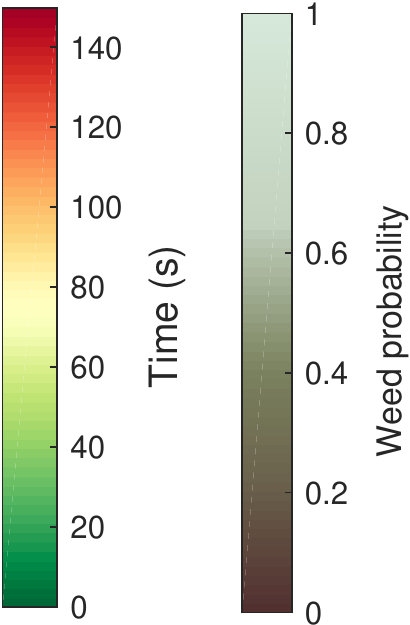}
   \caption{Recorded position over time for our \ac{IPP} algorithm with a random tag distribution. The final 
occupancy map is rendered. The `x' indicates the initial \ac{UAV} position. From here, 
an outward path is traveled (green-red) to explore the map edges. The lighter cells depict 9 out of 10 
successfully detected tags.}\label{F:experiment_path}
\end{figure}

~\figref{F:experiment_metrics} captures the variations in informative metrics for the experiments. The 
curves validate that our approach produces consistent real-time results independently of the target 
environment. For the random tag distribution, the traveled path is visualized 
in~\figref{F:experiment_path}\footnote{This experiment can be seen in the video attachment.}, depicting a plan 
similar to the simulated results (\figref{F:teaser}). The lighter map cells indicate locations of detected 
tags. As can be seen, the weakest performance for the random distribution in~\figref{F:experiment_metrics} 
(blue) arises due to inaccuracies in tag detection, leading to uncertainty in the associated cells.

\section{CONCLUSIONS AND FUTURE WORK}
In this work, we targeted the problem of planning informative paths for active classification. We presented 
an adaptive strategy that generates dynamically feasible paths in continuous 3D space for 
information-theoretic objectives. The approach was validated with one of the most important applications 
for precision agriculture, weed detection. An evaluation in simulation showed its advantages 
over a ``lawnmower'' coverage pattern and sampling-based \ac{IPP} algorithm in terms of informative metrics. 
We also demonstrated the effects of planning with different objectives and optimization strategies. Our 
experiments showed the framework running both in simulation and on a real multicopter platform.

Future work will target full field deployments with a real weed classifier. Interesting 
research directions involve considering different environment sizes and incorporating prior knowledge from 
previous scans.

\section*{ACKNOWLEDGMENT}
This work was funded by the European Community's Horizon 2020 programme under grant agreement no
644227-Flourish and from the Swiss State Secretariat for Education, Research and Innovation (SERI)
under contract number 15.0029. We would like to thank the ETH Crop Science
Group for providing the testing facilities and Rafael Jacinto G\'omez-Jordana Ma\~nas for his assistance in 
the experiments.

\bibliographystyle{IEEEtranN}
\footnotesize
\bibliography{references/IEEEabrv,references/2017-icra-popovic}

\end{document}

%% file: common_acronyms.tex
\begin{acronym}

\acro{2D}{two-dimensional}

\acro{3D}{three-dimensional}

\acro{AHRS}{attitude and heading reference system}

\acro{AUV}{autonomous underwater vehicle}

\acro{CPP}{Chinese Postman Problem}

\acro{DoF}{degree of freedom}
\acrodefplural{DoF}[DoFs]{degrees of freedom}

\acro{DVL}{Doppler velocity log}

\acro{FSM}{finite state machine}

\acro{IMU}{inertial measurement unit}

\acro{LBL}{Long Baseline}

\acro{MCM}{mine countermeasures}

\acro{MDP}{Markov decision process}
\acrodefplural{MDP}[MDPs]{Markov decision processes}

\acro{POMDP}{partially observable Markov decision process}
\acrodefplural{POMDP}[POMDPs]{partially observable Markov decision processes}

\acro{PRM}{Probabilistic Roadmap}
\acrodefplural{PRM}[PRM]{Probabilistic Roadmaps}

\acro{ROI}{region of interest}
\acrodefplural{ROI}[ROIs]{regions of interest}

\acro{ROS}{Robot Operating System}

\acro{ROV}{remotely operated vehicle}

\acro{RRT}{Rapidly-exploring Random Tree}
\acrodefplural{RRT}[RRTs]{Rapidly-exploring Random Trees}

\acro{SLAM}{simultaneous localization and mapping}

\acro{SSE}{sum of squared errors}

\acro{STOMP}{Stochastic Trajectory Optimization for Motion Planning}

\acro{TRN}{Terrain-Relative Navigation}

\acro{UAV}{unmanned aerial vehicle}

\acro{USBL}{Ultra-Short Baseline}

\acro{IPP}{informative path planning}

\acro{FoV}{field of view}
\acrodefplural{FoV}[FoVs]{fields of view}

\acro{CDF}{cumulative distribution function}

\acro{ML}{maximum likelihood}

\end{acronym}